%% file: main.tex
\documentclass{article}

\usepackage{hyperref}

\usepackage[table]{xcolor}
\usepackage[accepted]{icml2025}

\usepackage{amsmath}
\usepackage{amssymb}
\usepackage{mathtools}
\usepackage{amsthm}

\usepackage[capitalize,noabbrev]{cleveref}

\theoremstyle{plain}

\theoremstyle{definition}

\theoremstyle{remark}

\usepackage[disable,textsize=tiny]{todonotes}
\input{subtex/package.tex}

\icmltitlerunning{Normalizing Flows with Iterative Denoising}

\begin{document}

\twocolumn[
\icmltitle{Normalizing Flows with Iterative Denoising}

% It is OKAY to include author information, even for blind
% submissions: the style file will automatically remove it for you
% unless you've provided the [accepted] option to the icml2025
% package.

% List of affiliations: The first argument should be a (short)
% identifier you will use later to specify author affiliations
% Academic affiliations should list Department, University, City, Region, Country
% Industry affiliations should list Company, City, Region, Country

% You can specify symbols, otherwise they are numbered in order.
% Ideally, you should not use this facility. Affiliations will be numbered
% in order of appearance and this is the preferred way.
\icmlsetsymbol{equal}{*}

\begin{icmlauthorlist}
\icmlauthor{Tianrong Chen}{yyy}
\icmlauthor{Jiatao Gu}{yyy}
\icmlauthor{David Berthelot}{yyy}
\icmlauthor{Joshua Susskind}{yyy}
\icmlauthor{Shuangfei Zhai}{yyy}
\end{icmlauthorlist}

\icmlaffiliation{yyy}{Apple Machine Learning Research}
% \icmlaffiliation{comp}{Company Name, Location, Country}
% \icmlaffiliation{sch}{School of ZZZ, Institute of WWW, Location, Country}

\icmlcorrespondingauthor{Tianrong Chen}{tchen54@apple.com}
\icmlcorrespondingauthor{Shuangfei Zhai}{szhai@apple.com}

% You may provide any keywords that you
% find helpful for describing your paper; these are used to populate
% the "keywords" metadata in the PDF but will not be shown in the document
\icmlkeywords{Machine Learning, ICML}

\vskip 0.3in
]

% this must go after the closing bracket ] following \twocolumn[ ...

% This command actually creates the footnote in the first column
% listing the affiliations and the copyright notice.
% The command takes one argument, which is text to display at the start of the footnote.
% The \icmlEqualContribution command is standard text for equal contribution.
% Remove it (just {}) if you do not need this facility.

\printAffiliationsAndNotice{}  % leave blank if no need to mention equal contribution
% \printAffiliationsAndNotice{\icmlEqualContribution} % otherwise use the standard text.

\begin{abstract}
    % Normalizing Flows were once regarded as a promising class of likelihood-based generative models, yet their development has progressed more slowly compared with other modern approaches. Recent efforts such as TARFlow have revisited Autoregressive Normalizing Flows at scale by leveraging Transformer architectures. However, despite these advances, their empirical performance remains moderate and continues to lag behind the current state-of-the-art achieved by diffusion-based models and discrete autoregressive methods. 
    Normalizing Flows (NFs) are a classical family of likelihood-based methods that have received revived attention. Recent efforts such as TARFlow have shown that
    NFs are capable of achieving promising performance on image modeling tasks, making them viable alternatives to other methods such as diffusion models.
    In this work, we further advance the state of Normalizing Flow generative models by introducing iterative TARFlow (iTARFlow). Unlike diffusion models, iTARFlow maintains a fully end-to-end, likelihood-based objective during training. During sampling, it performs autoregressive generation followed by an iterative denoising procedure inspired by diffusion-style methods. Through extensive experiments, we show that iTARFlow achieves competitive performance across ImageNet resolutions of 64, 128, and 256 pixels, demonstrating its potential as a strong generative model and advancing the frontier of Normalizing Flows. In addition, we analyze the characteristic artifacts produced by iTARFlow, offering insights that may shed light on future improvements. Code is available at \url{https://github.com/apple/ml-itarflow}.
\end{abstract}

\input{subtex/intro.tex}
\input{subtex/prem.tex}
\input{subtex/method.tex}
\input{subtex/experiments.tex}

\input{subtex/conclusion.tex}
\bibliographystyle{icml2025}
\bibliography{example_paper}

%%%%%%%%%%%%%%%%%%%%%%%%%%%%%%%%%%%%%%%%%%%%%%%%%%%%%%%%%%%%%%%%%%%%%%%%%%%%%%%
%%%%%%%%%%%%%%%%%%%%%%%%%%%%%%%%%%%%%%%%%%%%%%%%%%%%%%%%%%%%%%%%%%%%%%%%%%%%%%%
% APPENDIX
%%%%%%%%%%%%%%%%%%%%%%%%%%%%%%%%%%%%%%%%%%%%%%%%%%%%%%%%%%%%%%%%%%%%%%%%%%%%%%%
%%%%%%%%%%%%%%%%%%%%%%%%%%%%%%%%%%%%%%%%%%%%%%%%%%%%%%%%%%%%%%%%%%%%%%%%%%%%%%%
\newpage
\appendix
\onecolumn
% \section{You \emph{can} have an appendix here.}

\end{document}

%% file: subtex/package.tex
\input{subtex/math_commands.tex}
\input{subtex/math_util.tex}
\usepackage{tabularx}
\usepackage{framed}
\usepackage{makecell}  % Add this in your preamble
\usepackage{multirow}
\newcommand\numberthis{\addtocounter{equation}{1}\tag{\theequation}}
\usepackage{cancel}

\usepackage{capt-of}
\usepackage{wrapfig}

\usepackage{xcolor}
\usepackage{color,soul}
\colorlet{color1}{green!50!black}
\colorlet{color2}{orange!95!black}
\colorlet{color3}{red!80!black}
\colorlet{color4}{red!65!black}
\colorlet{color5}{blue!75!green}
\colorlet{blueee}{blue!50!black}

\definecolor{amaranth}{rgb}{0.9, 0.17, 0.31}

\usepackage{footnote}
\let\oldsqrt\sqrt
\def\sqrt{\mathpalette\DHLhksqrt}
\def\DHLhksqrt#1#2{\setbox0=\hbox{$#1\oldsqrt{#2\,}$}\dimen0=\ht0
\advance\dimen0-0.2\ht0
\setbox2=\hbox{\vrule height\ht0 depth -\dimen0}%
{\box0\lower0.4pt\box2}}

\usepackage{enumitem}
\usepackage{subfig}
\usepackage{arydshln}

\usepackage{algorithm}
\usepackage{algorithmic}

\usepackage{booktabs}       % professional-quality tables

\usepackage{tikz}
\usepackage{empheq}

\usepackage{tablefootnote}
\usepackage{textcomp}
\usepackage{colortbl}
\usepackage{aligned-overset}

%% file: subtex/math_commands.tex
\usepackage{amsmath,amsfonts,bm}

\def\eqref#1{(\ref{#1})}
\def\1{\bm{1}}

\def\rd{{\textnormal{d}}}

\def\rmI{{\mathbf{I}}}

\def\vf{{\bm{f}}}

\def\vx{{\bm{x}}}

\def\vz{{\bm{z}}}

\DeclareMathAlphabet{\mathsfit}{\encodingdefault}{\sfdefault}{m}{sl}
\SetMathAlphabet{\mathsfit}{bold}{\encodingdefault}{\sfdefault}{bx}{n}

\newcommand{\pdata}{p_{\rm{data}}}
\newcommand{\tmin}{t_{\rm{min}}}
\newcommand{\tmax}{t_{\rm{max}}}
\newcommand{\pprior}{p_{\rm{prior}}}

%% file: subtex/intro.tex
\begin{figure*}[t]
    \centering
    \includegraphics[width=\textwidth]{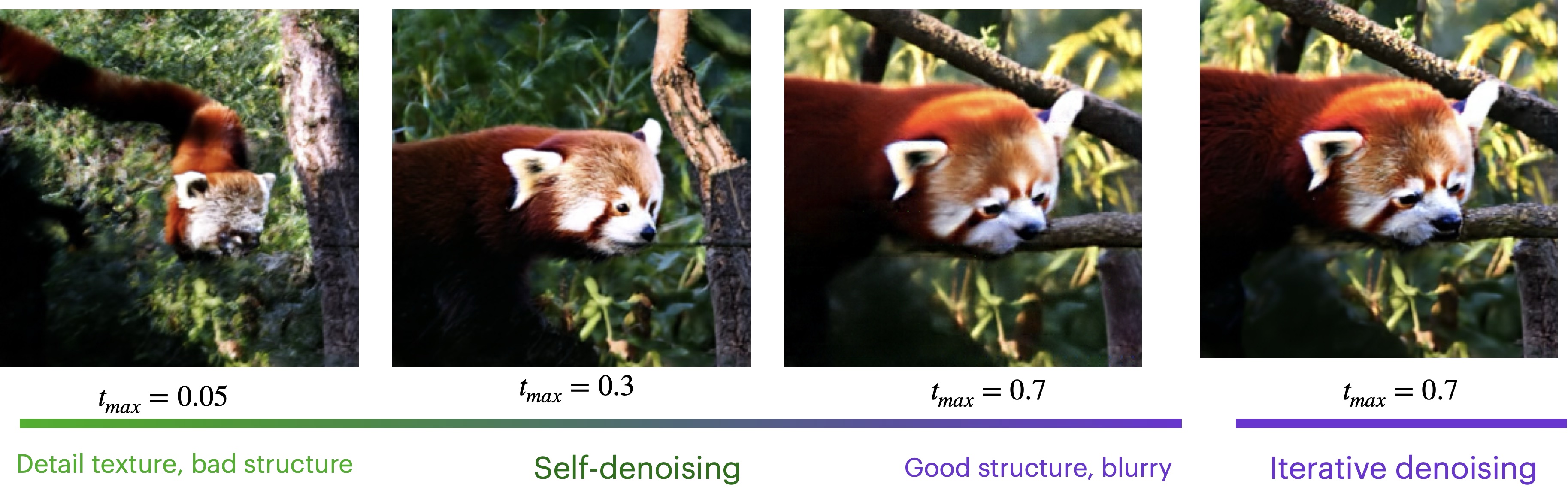}
    \caption{\textbf{Demonstration of noise dilemma:} When the maximum noise level $\tmax$ used during training is too small, the model tends to generate images with rich local textures but poor global structure. Conversely, when the maximum noise level $\tmax$ is large, the model generates samples with accurate global structure but noticeably blurred fine details and visible artifacts (zoom in for details), even after self-denoising. Interestingly, by applying iterative denoising, the model is able to leverage the strengths of both regimes, yielding images that preserve both structural coherence and fine-grained texture.}
    \label{fig:noise_dilimma}
\end{figure*}
\begin{figure*}[t]  % use [t] or [b] to control placement
    \centering
    \includegraphics[width=\textwidth]{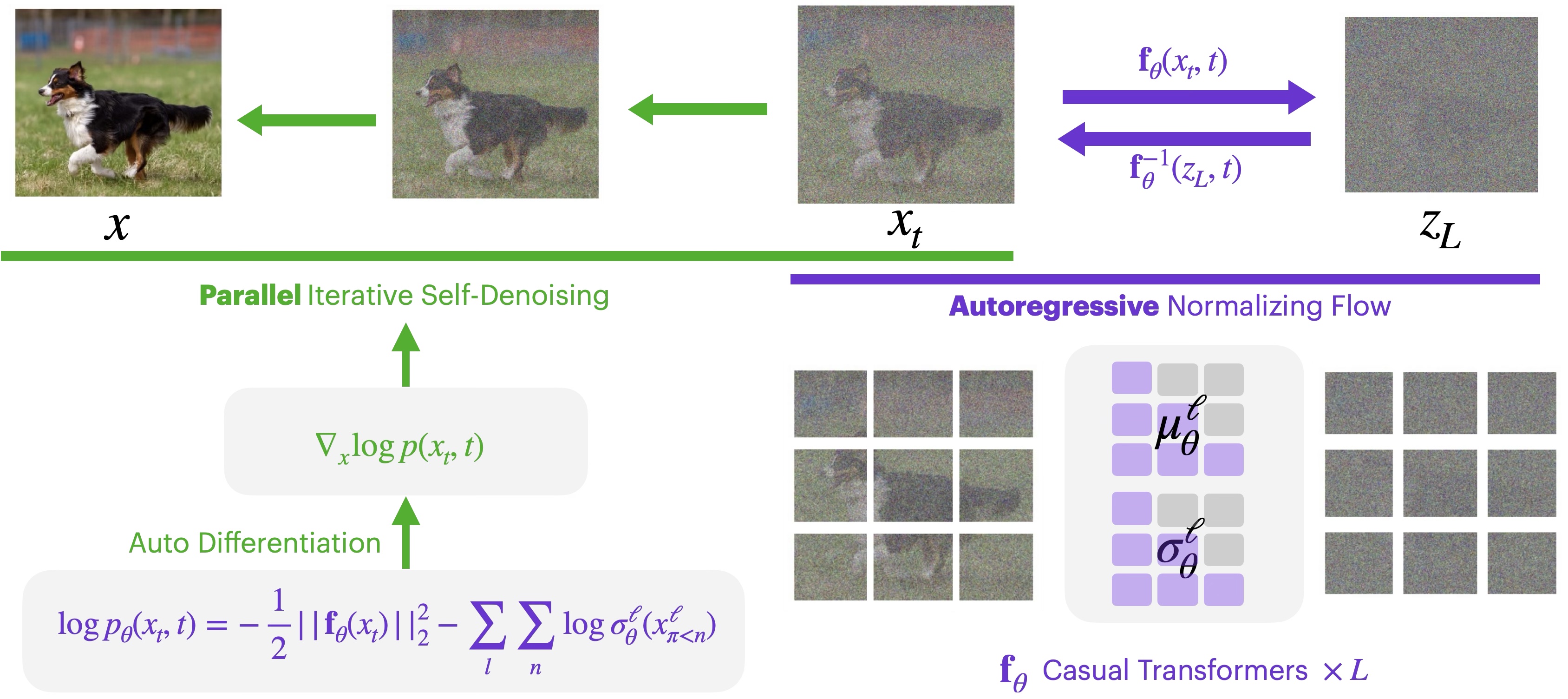}
    \caption{\textbf{Demonstration of iTARFlow.} During training, we optimize a TARFlow across a range of noise levels using a shared network, analogous to diffusion models. The TARFlow is an invertible, causal Transformer-based NF
    composed of $L$ stacked causal Transformer blocks. During sampling, the model first
    generates a noisy sample $\vx_t$ and then performs iterative denoising by
    leveraging automatic differentiation of the parameterized log-likelihood across
    noise levels.}
    \label{fig:algo_demo}
  \end{figure*}

\section{Introduction}
In recent years, advances in image generation have been shaped primarily by two dominant modelling paradigms. The first is the class of diffusion models~\cite{ho2020denoising,rombach2022high,peebles2023scalable,esser2024scaling}, which synthesize images through continuous-space denoising dynamics and are capable of producing state-of-the-art visual quality. Despite their success, diffusion models suffer from two fundamental limitations. They require dozens to hundreds of iterative denoising steps at inference time, resulting in substantial computational overhead. More critically, their scaling behavior remains far less predictable~\cite{liang2024scaling} compared to Large Language Models, making it unclear whether diffusion-based systems can benefit from the same scaling laws \cite{kaplan2020scaling} that have driven recent breakthroughs in language models. In contrast, the rapid progress of Large Language Models has demonstrated the remarkable scalability and capability of autoregressive architectures. Motivated by these successes, the second family of image generative models is discrete autoregressive models~\cite{yu2022scaling,sun2024autoregressive,tian2024visual}. They represent data as sequences of discrete tokens and leverage Transformer architectures \cite{vaswani2017attention} to model data distributions. These models offer fast sampling and well-understood scaling properties, but quantization inevitably introduces an information bottleneck that can degrade fidelity. This naturally prompts the question of whether one can combine the scalability benefits of autoregressive modeling with the expressiveness of continuous-valued image representations. Early attempts in this direction explored autoregressive models directly in continuous pixel space, beginning with convolutional models such as PixelRNN and PixelCNN~\cite{van2016conditional,van2016pixel} and later extending to Transformer-based variants such as Image Transformer~\cite{parmar2018image} and iGPT~\cite{chen2020generative}. While these methods avoid quantization, they operate on pixel-level sequences that are extremely long, making both training and sampling slow and causing difficulties in modeling long spatial structure.

These observations motivate the search for a generative model framework that preserves the autoregressive structure shown to scale effectively in language models, while also retaining the advantages of continuous representations crucial for high-fidelity visual synthesis. Normalizing flows (NFs)~\cite{tabak2010density,rezende2015variational,dinh2016density} offer a possibility toward such a model class. NFs define an invertible mapping $\bf:\mathbb{R}^{D}\rightarrow\mathbb{R}^D$ that transforms the data distribution $\pdata$ into a simple prior, enabling exact maximum-likelihood training analogous to discrete autoregressive language models. Subsequent developments introduced Autoregressive Flow (AF) architectures~\cite{papamakarios2017masked,kingma2016improving}, providing a mechanism to incorporate autoregressive structure into NFs. 

Recently, TARFlow~\cite{zhai2024normalizing} and its subsequent works~\cite{gu2025starflow,zhang2025flexible,guend} demonstrate that NFs are capable of producing competitive performance on image/video tasks. This is achieved by leveraging a powerful Transformer \cite{vaswani2017attention} based AF architecture which enjoys stable training and excellent scalability. 

A subtle but critical component of TARFlow is the use of additive Gaussian noise, instead of the small dequantization noise used in prior NF instances, to the input. Likelihood is consequently over the noisy data distribution, which is shown to greatly improve the generalization/sampling ability of the model. During sampling time, a score-based denoising step is employed to clean up the noise portion in the samples, where the score is computed directly as the derivative of the loss function w.r.t.\ its input.
% Due to the inherent properties of NFs, input data quantization such as convolving the data with a small Gaussian kernel is necessary to prevent unbounded likelihoods and ensure stable training. As a consequence, After training, a NF model can only generate noisy data, since all training examples presented to the network are noisy observations of the data. Because the model directly parameterizes the data likelihood, its score function can be obtained automatically via differentiation of the log-likelihood with respect to the input. This score estimation enables us to denoise the noisy samples and recover clean images. 

A central question then arises: how much noise should be injected into the training data to achieve both stable optimization and high visual quality after denoising? In this work, we conduct a systematic study of this problem and reveal a phenomenon we refer to as the noise dilemma. Specifically, when the training noise level is too small, the denoised samples exhibit excessively rich textures and appear artificially detailed (see the left side of Fig.~\ref{fig:noise_dilimma}). In contrast, when the training noise is large, the subsequent self-denoising process produces overly smooth, blurry images (right side of Fig.~\ref{fig:noise_dilimma}) due to the behavior predicted by Tweedie’s lemma~\cite{robbins1992empirical}.

Motivated by these findings, we propose to train the Transformer Autoregressive Flow (TARFlow; \citet{zhai2024normalizing}) across a range of noise levels and to perform iterative score-based denoising at inference time. We refer to this method as iterative TARFlow (iTARFlow). This strategy allows TARFlow to generate highly noisy but globally coherent images using large noise levels in an autoregressive fashion, and then refine them in parallel using scores obtained from the likelihood parameterization. The key enabling factor is that iTARFlow is trained explicitly across multiple noise scales. Empirically, iTARFlow achieves substantial improvements over previous TARFlow-based models and significantly narrows the performance gap between normalizing flows, diffusion models, and discrete autoregressive generative models. Our contributions are summarized as follows:
\setlength{\leftmargini}{1em}
\begin{itemize}
    \item We identify and characterize the noise dilemma in NF training, showing that higher noise levels surprisingly help produce images with better global structure. Building on this observation, we propose a multi-noise training scheme for TARFlow and develop an inference pipeline that generates high-noise samples and subsequently denoises them through iterative, likelihood-induced score estimation.
    \item We demonstrate that iTARFlow outperforms prior TARFlow variants and further closes the gap between NFs and current state-of-the-art generative models. These results highlight the strong potential of our approach for future NFs generative modeling.
    \item Although our method achieves competitive quantitative performance, we also report in Sec.~\ref{sec:failure} that two specific types of failure cases appear among the generated samples used for the metric evaluation. These observations suggest that the model’s performance could be further improved by addressing these remaining issues in the future.
\end{itemize}

%% file: subtex/prem.tex
\begin{figure*}[t]  % use [t] or [b] to control placement
    \centering
    \includegraphics[width=\textwidth]{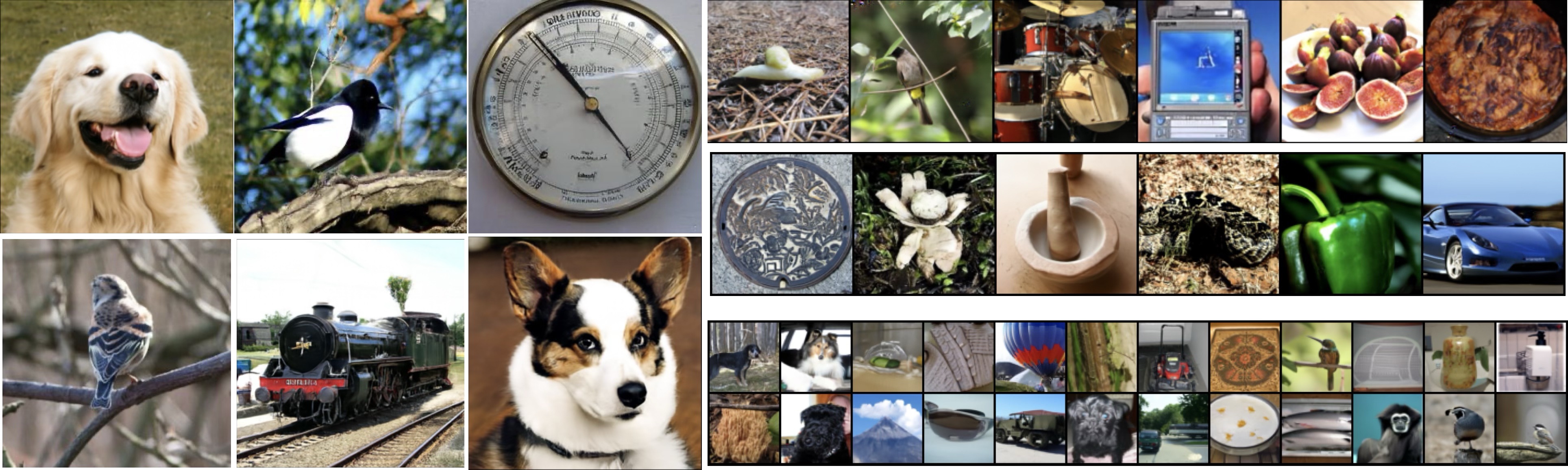}
    \caption{Samples from iTARFlow in pixel space. Left to right, top to bottom: ImageNet 256, 128, and 64 resolution with patch sizes 8, 4, and 2, respectively.}
    \label{fig:teaser}
  \end{figure*}
\section{Preliminary}
\begin{table}[h]
    \caption{Notations used in this paper}
    \centering
    \begin{tabular}{ll}
    \hline
    Notation & Meaning \\
    \hline
    $L$ & number of causal transformer blocks\\
    $\ell$ & $\ell$-th causal transformer block\\
    $\vf$ & stacked causal transformer blocks\\
    $N,n$ & Total number of tokens, $n$ is index\\
    $y$ & class conditioning\\
    $p_{\theta}$ & model distribution \\
    $\pdata$ & empirical data distribution\\
    $\pprior$ & prior distribution\\
    \hline
    \end{tabular}
    \end{table}
\subsection{Normalizing Flows}
% \paragraph{Notations}:
% \subsection{Preliminaries: Autoregressive Normalizing Flows}

Normalizing Flows are a class of generative models that learn an invertible 
transformation $\vf: \mathbb{R}^D \rightarrow \mathbb{R}^D$ that maps the 
data distribution $\pdata$ to a simple, tractable prior such as the standard 
Gaussian. Given a sample $\vx \sim \pdata$, its transformed representation 
$\vz = \vf(\vx)$ is encouraged to follow the prior distribution 
$p_{\mathrm{prior}}$. By the change-of-variables formula~\cite{rezende2015variational}, 
the data log-likelihood is:
\begin{align}\label{eq:mle}
    \log p_{\theta}(\vx)
    = \log p_{\mathrm{prior}}(\vf(\vx))
      + \log \left|
        \det\left(
            \frac{\partial \vf(\vx)}{\partial \vx}
        \right)
      \right|.
\end{align}
Maximizing this likelihood corresponds to learning a transformation that pushes 
$\pdata$ into a simple prior while accounting for the Jacobian volume change. 
To generate new samples, one draws $\vz \sim p_{\mathrm{prior}}$ and applies 
the inverse mapping $\vx = \vf^{-1}(\vz)$.

\paragraph{Autoregressive Normalizing Flows.}
To ensure tractable Jacobian determinants and efficient inversion, 
Autoregressive Normalizing Flows~\cite{kingma2016improved,papamakarios2017masked} 
impose an autoregressive structure on the transformation. Recent developments 
in this direction~\cite{zhai2024normalizing,zhang2025flexible,gu2025starflow} 
implement the flow as a composition of causal Transformer blocks:
\begin{align*}
    \vf
    = f^{L} \circ f^{L-1} \circ \cdots \circ f^1.
\end{align*}
Each block $f^{\ell}$ is a causal Transformer that outputs the affine parameters 
$\mu^{\ell}_{\theta}$ and $\sigma^{\ell}_{\theta}$ for each token. For example, consider an image $\vx \in \mathbb{R}^D$ tokenized into patches of size 
$p \times p$ with $c$ channels, producing a sequence of $N$ tokens. For token
index $n < N$, the forward and inverse transformations of a single flow layer
$f_{\ell}$ are:
\begin{align*}
    &\textbf{forward:}
    && \vz^{\ell+1}_n
      = \frac{\vx^{\ell}_n - \mu^{\ell}_{\theta}(\vx^{\ell}_{<n})}
             {\sigma^{\ell}_{\theta}(\vx^{\ell}_{<n})},
      \numberthis \\
    &\textbf{inverse:}
    && \vx^{\ell}_n
      = \mu^{\ell}_{\theta}(\vx^{\ell}_{<n})
        + \sigma^{\ell}_{\theta}(\vx^{\ell}_{<n})\, \vz^{\ell+1}_n,
    \numberthis \label{eq:flow_reverse}
\end{align*}where the causal conditioning structure on $\vx^{\ell}_{<n}$ ensures tractable sampling and 
log-determinant evaluation.

To enhance expressiveness across successive flow layers, a permutation $\pi$ is applied to the token sequence between adjacent layers. Formally, $\pi : \{1,\dots,N\} \to \{1,\dots,N\}$ is a bijection acting on token indices. In prior work, the chosen permutation is the simple flip mapping $\pi(n) = N - n + 1$. Under this transformation, the autoregressive conditioning for the $\ell$-th flow layer changes from $x^{\ell}_{<n}$ to
\[
    \vx^{\ell}_{\pi < n}
    = \big( \vx^{\ell}_{N}, \vx^{\ell}_{N-1}, \dots, \vx^{\ell}_{N-n+2} \big),
\]
allowing the model to capture dependencies from both directions of the token sequence and ensuring universal approximation capability (Prop.~1 in~\cite{gu2025starflow}). Because permutations correspond to volume-preserving linear maps with $|\det P_{\pi}| = 1$, they contribute no additional term to the log-likelihood and admit trivial inverses. Consequently, permutation layers increase expressive power without introducing extra computational cost. With these design 
choices, the training objective remains simple:
\begin{align*}
    &\max_{\theta}\, \mathbb{E}_{\vx \sim \pdata}\log p_{\theta}(\vx;\theta)\\
    &= -\frac{1}{2}\left\|\vf_{\theta}(\vx)\right\|_2^2
       - \sum_{\ell=1}^{L}\sum_{n=1}^{N} 
         \log \sigma^{\ell}_{\theta}\!\left(\vx^{\ell}_{\pi < n}\right),
    \numberthis \label{eq:logp}
\end{align*}

\subsection{Noise Dilemma}\label{sec:noise_dilemma}
% \paragraph{Dequantization and Self-Denoising.}
`Dequantization' noise~\cite{dinh2016density,ho2019flow++} was the standard procedure for converting discrete/quantized data to continuous inputs used in prior NFs. However, in TARFlow~\cite{zhai2024normalizing}, it was shown that there is great benefit in using additive Gaussian noise with tunable magnitude, which significantly improves the model's sampling capabilities. Formally, this is defined as
\[
\vx_t = \vx + t \boldsymbol{\epsilon}, \quad \epsilon \sim \mathcal{N}(0, \mathbf{I}).
\]
Consequently, the normalizing flow model learns to represent the data distribution convolved with Gaussian noise, implying that the generated samples are inherently noisy. To recover clean samples, a \emph{self-denoiser} was proposed in~\cite{zhai2024normalizing}. Specifically, since the model is trained to maximize $\log p_{\theta}(x_t)$, one can obtain the score function $\nabla_{x} \log p(x_t)$ via automatic differentiation~\cite{paszke2019pytorch} at noise level $t$. The optimal denoiser follows Tweedie’s lemma \cite{robbins1992empirical}:
\begin{align}
    \hat{\vx} := \vx_t + t^2 \nabla \log p_{\theta}(\vx_t).
\end{align}

However, NFs face the \emph{noise dilemma}, that is, the challenge of selecting an appropriate amount of injected noise. When the injected noise scale $t$ is small, the model tends to generate overly detailed, high-frequency images, as shown on the left-hand side of Fig.~\ref{fig:noise_dilimma}; furthermore, training tends to be unstable due to the unbounded nature of log-likelihood training with a small amount of noise added. Conversely, when a higher noise scale is applied, the model captures better global structures (right-hand side of Fig.~\ref{fig:noise_dilimma}) but produces blurrier results. This phenomenon also aligns with the theoretical upper bound of the optimal denoiser given by Tweedie’s lemma. Theoretically, the posterior mean corresponds to a smoothed reconstruction, which naturally becomes blurrier as the noise level increases.

To partially alleviate this issue, \citet{gu2025starflow} adopts a moderate noise regime for modeling the latent distribution, under which the self-denoiser becomes less effective. Instead, a fine-tuned VAE is utilized to map noisy latent codes back to the pixel space. \citet{guend} replaces the fine-tuned VAE with a jointly trained score model for denoising. Later, \citet{zheng2025farmer} proposes curriculum training, wherein the NF is initially trained with a moderate amount of noise, and the noise level is gradually reduced throughout the training process. Similar to curriculum training, \citet{kim2020softflow} amortize training across multiple noise levels using a noise-scale conditioning mechanism, and ultimately generate samples at the zero-noise limit. This corresponds to the left side of Fig.~\ref{fig:noise_dilimma}.

%% file: subtex/method.tex
\section{Method}
\begin{figure}[t]
    \centering
    \includegraphics[width=\columnwidth]{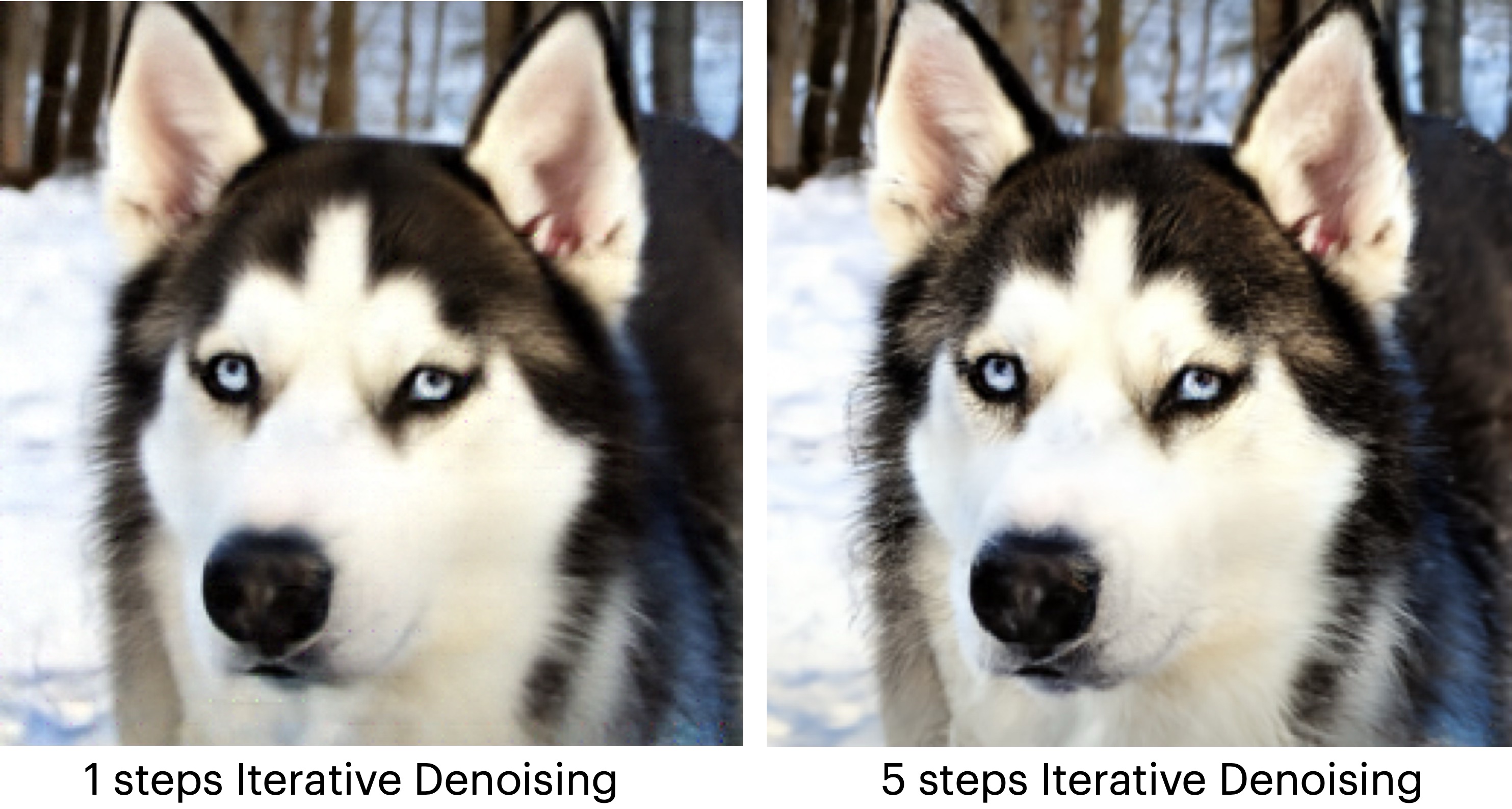}
    \caption{Demonstration of how the number of denoising steps affects visual quality (best viewed when zoomed in). With a single step, the generated image is blurry and contains artifacts, whereas using just five steps already yields visually high-quality results.}\label{fig:iter_ablation}
    \vspace{-8pt}
\end{figure}

% \begin{figure}[t]
%     \centering
%     \includegraphics[width=\columnwidth]{figs/tmax_ablation.png}
%     \caption{FID vs.\ training epoch.}
%     \label{fig:tmax_ablation}
% \end{figure}

\begin{figure}[t]
    \centering
    \includegraphics[width=\columnwidth]{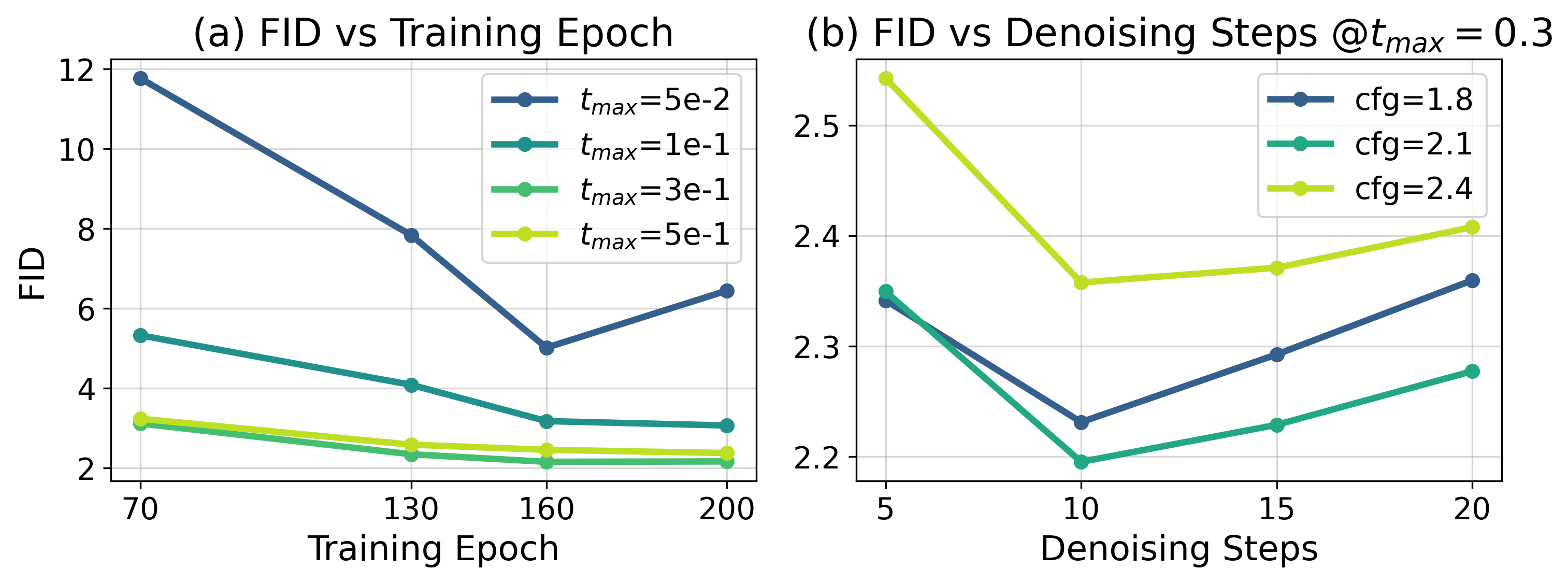}
    \caption{\textbf{(a)} Ablation study of the choice of $\tmax$ given $\tmin=0.01$ over training epochs. \textbf{(b)} Ablation study of the number of denoising steps used in iTARFlow. Since the normalizing flow already produces samples with relatively small noise, the number of steps required to obtain a clean image is typically not large.}
    \vspace{-20pt}
    \label{fig:ablation_tmax_step}
\end{figure}

\begin{figure}[t]
    \centering
    \includegraphics[width=0.9\columnwidth]{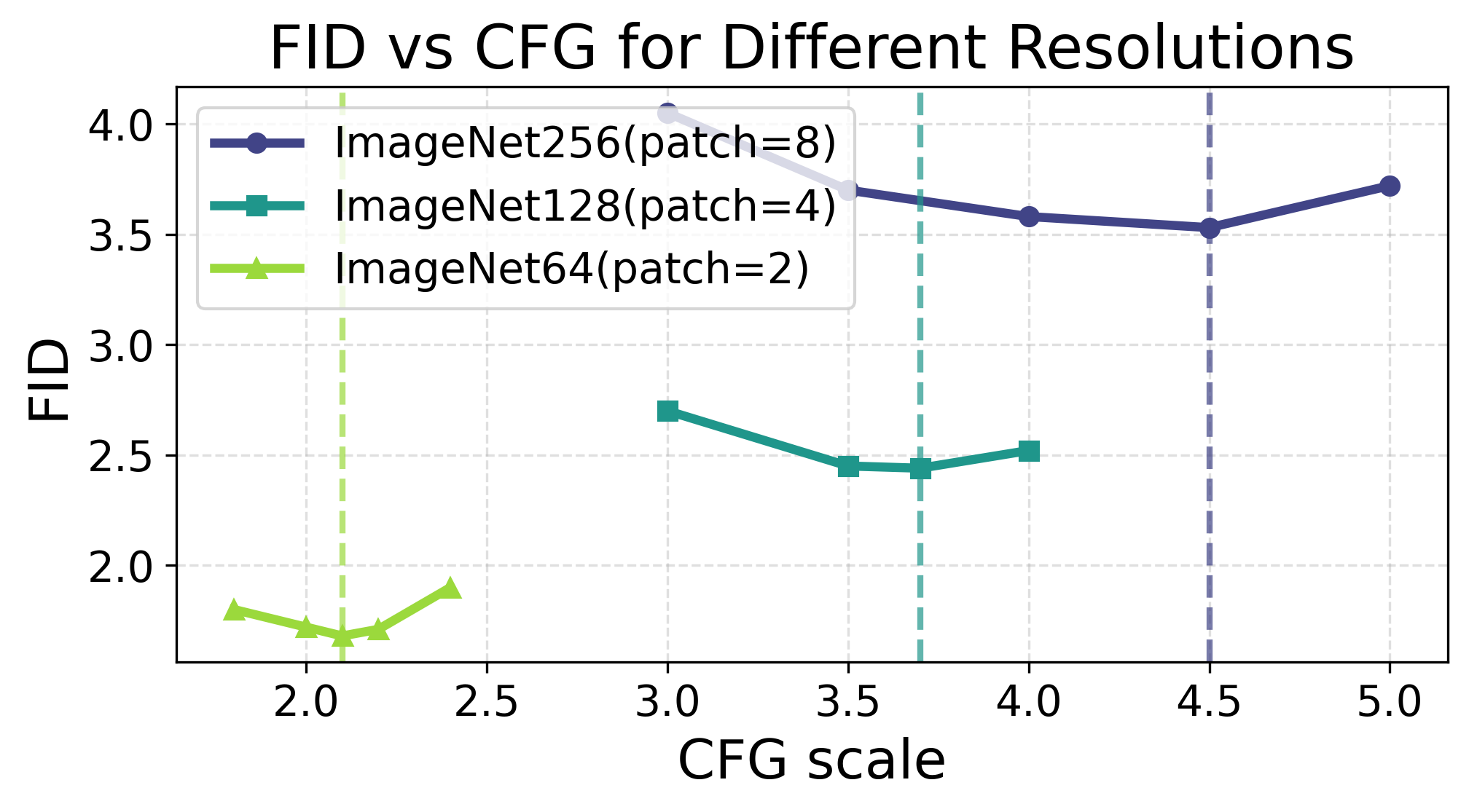}
    \caption{Ablation study of patch size across ImageNet resolutions. The results show that the optimal CFG scale differs substantially depending on the patch size.}
    \label{fig:cfg_ablation}
\end{figure}

This work introduces a simple yet effective framework that augments NFs with an iterative denoising procedure. The central idea is to train a likelihood model $\log p_{\theta}(x_t, t)$ over a continuum of noise levels $t \in [t_{\min}, t_{\max}]$. After the NF model produces an intermediate noisy sample $x_t$, the model refines it toward a clean image through iterative denoising, in a manner analogous to diffusion-based generation. Our Normalizing Flow follows the architecture used in TARFlow~\cite{zhai2024normalizing}; the backbone is implemented using Transformer blocks. Thus the resulting method is referred to as iterative TARFlow (iTARFlow).

\subsection{Training}\label{sec:train}
To learn score functions at multiple noise levels, the model must be trained across a range of noise intensities. Our goal is to achieve this with minimal deviation from standard NF training. The most direct strategy is to amortize the likelihood objective over $t$, enabling the model to jointly learn $\log p_{\theta}(x_t, t)$ for all noise levels. Specifically, we optimize the iTARFlow using the following objective function:
\begin{align*}
    \min_{\theta}\mathbb{E}_{\vx_t,t}&\left[\min_{\theta} \gamma_t\log p_{\theta}(\vx_t,t)\right] \quad \text{s.t.} \ \ \vx_t=\vx+t \boldsymbol{\epsilon} \\
    &\boldsymbol{\epsilon}\sim \mathcal{N}(0,\rmI), \ \  t \sim \mathcal{U} [\tmin,\tmax]
\end{align*}
where $\log p_{\theta}(\cdot)$ corresponds to the original TARFlow objective (Eq.~\ref{eq:mle}) with an additional conditioning on $t$. This formulation introduces three hyperparameters that require careful selection. The first is the choice of the training noise range $\tmin$ and $\tmax$. If $\tmin$ is too small, optimization becomes unstable due to the unbounded behavior of the negative log-likelihood near the data manifold. Conversely, setting $\tmin$ too large reduces the achievable sampling quality, as the final denoising step must begin from $\tmin$ rather than approaching the clean-data limit. As discussed in Sec.~\ref{sec:noise_dilemma}, the Tweedie Lemma further demonstrates that excessive noise scale leads to oversmoothing and blur. Based on these considerations, we set $\tmin = 0.01$, which yields stable optimization while preserving high-fidelity reconstructions.

Unfortunately, no theoretical principle dictates the optimal choice of the maximum noise scale $\tmax$. We therefore conduct an empirical ablation to determine a suitable value for $\tmax$. As illustrated in Fig.~\ref{fig:ablation_tmax_step}, $\tmax$ has a substantial impact on performance, with larger values generally yielding better results. For higher-dimensional datasets, we increase $\tmax$ accordingly. The specific configurations used in our experiments are provided in Sec.\ref{sec:exp}. Finally, following diffusion-model practice, we incorporate a noise scale dependent reweighting term $\gamma_t$ in the training objective. Our observations indicate that different choices of $\gamma_t$ perform similarly; thus, for simplicity and numerical stability, we adopt the simple choice $\gamma_t := t$.

\subsection{Sampling}

The sampling procedure is divided into two phases, both using the \emph{same trained iTARFlow model}. The first phase autoregressively constructs the noisy image $\vx_t$ patch by patch as dictated by the nature of TARFlow (Eq.~\ref{eq:flow_reverse}), while the second phase subsequently decodes it through a fully parallel, image-level denoising process.
\vspace{-0.2cm}
\paragraph{Autoregressive Normalizing Flow Generation}  
Once the iTARFlow has been trained to map Gaussian noise to the noisy data distribution $x_t$, sampling begins by drawing from the Gaussian prior and generating an image at noise level $\tmax$ using Eq.~\ref{eq:flow_reverse}. This stage proceeds in an autoregressive manner, producing patches sequentially. Following TARFlow and STARFlow, we heuristically apply patch-wise Classifier-Free Guidance (CFG) \cite{ho2022classifier,zhai2024normalizing,gu2025starflow} only at the first ($\ell=1$) causal Transformer layer for each patch during inference. We additionally observe that the optimal CFG scale depends on the patch size: larger patches usually benefit from stronger guidance. The corresponding ablation study is shown in Fig.~\ref{fig:cfg_ablation}.
\vspace{-0.1cm}
\paragraph{Parallel Denoising}  
After obtaining the noisy image $\vx_{\tmax}$ from the autoregressive generation, we apply a global denoising procedure to remove the remaining noise. Since our model parameterizes the TARFlow at each noise level through $\log p_{\theta}(\vx_t, t)$, the corresponding score function $\nabla_\vx \log p_{\theta}(\vx_t, t)$ can be computed directly via automatic differentiation. Unlike the autoregressive sampling required in the first phase, the denoising stage involves only forward evaluations of iTARFlow across different noise levels. This procedure is fully parallelizable and compatible with automatic differentiation, making it computationally efficient. The denoising process follows a probabilistic ODE which is closely related to Flow Matching~\cite{lipman2022flow} and DDIM~\cite{song2020denoising}. In this work, we adopt the following ODE formulation:
\begin{align*}
    \rd \vx_t
    = -\, t\, \nabla_\vx \log p_{\theta}(\vx_t, t)\, \mathrm{d}t,
    \qquad t \in [\tmin, \tmax].
\end{align*}
An ablation study on the number of ODE steps under linear discretization is provided in Fig.~\ref{fig:ablation_tmax_step}, with qualitative visualizations shown in Fig.~\ref{fig:iter_ablation}. These results indicate that as few as five steps are sufficient to produce visually high-quality samples, while ten steps are adequate to reach optimal quantitative performance, measured by Fr\'echet Inception Distance (FID)~\cite{heusel2017ttur}. During the denoising phase, we do not apply CFG, as it does not provide noticeable improvements and increases computational cost.

%% file: subtex/experiments.tex
\begin{table}[t]
  \centering
  \scriptsize
  \renewcommand{\arraystretch}{0.95}
  \caption{Overview of used network architectures}\label{Table:network_arch}
  \resizebox{\columnwidth}{!}{%
  \begin{tabular}{lccc}
  \toprule
  \textbf{Model} & 
  \makecell[c]{num. Attention Layer\\@ each Transformer block} 
  & patch size & Channel Size \\
  \midrule
  Small (S) Model & [2,2,2,12] & Resolution/32 & 1280 \\
  Big (B) Model & [4,4,4,24] & Resolution/32 & 1280 \\
  Large (L) Model & [4,4,4,24] & Resolution/32 & 1600 \\
  Extra Large (XL) Model & [4,4,4,24] & Resolution/32 & 2176 \\
  \bottomrule
  \end{tabular}%
  }
\end{table}

\begin{table}[t]
  \centering
  \scriptsize
  \renewcommand{\arraystretch}{0.95}
  \caption{Comparison of models on ImageNet-64. }\label{tab:imagenet64}
  \resizebox{\columnwidth}{!}{%
  \begin{tabular}{lcc}
  \toprule
  \textbf{Model} & \textbf{FID} $\downarrow$ & \textbf{\# Param.} \\
  \midrule
  \multicolumn{3}{l}{\textbf{Diffusion Models}} \\
  % \midrule
  EDM-SDE (511 NFE) \cite{karras2022elucidating} & 1.55 & 300M \\
  EDM-ODE (79 NFE) \cite{karras2022elucidating} & 2.36 & 300M \\
  iDDPM \cite{nichol2021improved} & 2.92 & 300M \\
  ADM(dropout) \cite{nichol2021improved} & 2.09 & 554M \\
  \midrule
  \multicolumn{3}{l}{\textbf{Consistency Models}} \\
  CD(LPIPS) \cite{song2023consistency} & 4.70 & -- \\
  iCT-deep \cite{song2023improved} & 3.25 & -- \\
  \midrule
  \multicolumn{3}{l}{\textbf{Normalizing Flow}} \\
  % \midrule
  TARFlow \cite{zhai2024normalizing} & 2.66 & 880M \\
  \rowcolor{gray!15}
  iTARFlow-S (ours) & 2.05 & 350M \\
  \rowcolor{gray!15}
  iTARFlow-B (ours) & 1.68 & 770M \\
  \bottomrule
  \end{tabular}%
  }
\end{table}
\begin{table}[t]
  \centering
  \scriptsize
  \renewcommand{\arraystretch}{0.95}
  \caption{Comparison of models on ImageNet-128.}\label{tab:imagenet128}
  \resizebox{\columnwidth}{!}{%
  \begin{tabular}{lcc}
  \toprule
  \textbf{Model} & \textbf{FID} $\downarrow$ & \textbf{\# Param.} \\
  % \midrule
  \multicolumn{3}{l}{\textbf{Diffusion Models}} \\
  \midrule
  ADM-G (511 NFE) \cite{nichol2021improved} & 2.97 & 554M \\
  CDM \cite{ho2022cascaded} & 3.52 & -- \\
  Simple Diff \cite{hoogeboom2023simple} & 1.94 & 2B \\
  RIN \cite{jabri2022scalable} & 2.75 & 410M \\
  \midrule
  \multicolumn{3}{l}{\textbf{Normalizing Flow}} \\
  % \midrule
  TARFlow \cite{zhai2024normalizing} & 5.03 & 1.3B \\
  \rowcolor{gray!15}
  iTARFlow-L (ours) & 2.44 & 1.2B \\
  \bottomrule
  \end{tabular}%
  }
\end{table}
\begin{table}[t]
  \centering
  \scriptsize
  \renewcommand{\arraystretch}{0.95}
  \caption{Comparison of models on ImageNet-256 generation in pixel and latent space.}\label{tab:imagenet256}
  \resizebox{\columnwidth}{!}{%
  \begin{tabular}{lcc}
  \toprule
  \textbf{Model} & \textbf{FID} $\downarrow$ & \textbf{\# Param.} \\
  \midrule
  \multicolumn{3}{l}{\textbf{\underline{Latent} Diffusion Models}} \\
  LDM-4\cite{rombach2022high} & 3.6 & 400M+86M \\
  DiT-XL \cite{peebles2023scalable} & 2.27 & 675M+86M \\
  SiT-XL \cite{ma2024sit} & 2.06 & 675M+86M \\
  REPA \cite{yu2024representation} & 1.94 &675M+86M \\
  \midrule
  \multicolumn{3}{l}{\textbf{\underline{Latent} Autoregressive Models}} \\
  GIVT\cite{tschannen2024givt} &  2.59& 1.67B+53M\\
  MAR-AR \cite{li2024autoregressive} & 4.69 & 479M+66M \\
  MAR-L \cite{li2024autoregressive} & 1.78 &479M+66M \\
  \midrule
  \multicolumn{3}{l}{\textbf{\underline{Latent} Normalizing Flow}} \\
  % \midrule
  STARFLOW \cite{gu2025starflow} & 2.40 & 1.4B+86M \\
  \rowcolor{gray!15}
  iTARFlow-B (ours) & 2.32 & 770M+86M \\
  \midrule
  \midrule
  \multicolumn{3}{l}{\textbf{\underline{Pixel} Diffusion Model}} \\
  ADM\cite{dhariwal2021diffusion} &  4.59& 554M \\
  CDM \cite{ho2022cascaded} & 4.88 & -\\
  Simple-Diff (Unet) \cite{hoogeboom2023simple} & 3.76 &- \\
  Simple-Diff (UViT) \cite{hoogeboom2023simple} & 2.77 &2B \\
  PixNerd-XL \cite{wang2025pixnerd}& 1.93 &700M\\
  SiD2 patch 1 \cite{hoogeboom2024simpler} & 1.38 &-\\
  \midrule
  \multicolumn{3}{l}{\textbf{\underline{Pixel} Autoregressive Model}} \\
  FractalMAR-H\cite{li2025fractal} &  6.15& 844M \\
  \midrule
  \multicolumn{3}{l}{\textbf{\underline{Pixel} Normalizing Flow}} \\
  JetFormer \cite{tschannen2024jetformer} & 6.64 & 2.8B\\
  FARMER-Patch8 \cite{zheng2025farmer} & 3.60 &1.9B \\
  TARFlow\cite{zhai2024normalizing} &  5.56& 1.3B \\
  STARFlow\cite{gu2025starflow} &  4.69& 1.4B \\
  \rowcolor{gray!15}
  iTARFlow-XL(ours)&  3.32& 2.2B \\
  \bottomrule
  \end{tabular}%
  }
\end{table}

\section{Experiments}\label{sec:exp}

In this section, we evaluate the performance of iTARFlow on ImageNet at resolutions of 64, 128, and 256 pixels~\cite{deng2009imagenet}, and compare iTARFlow against baselines in pixel and latent space settings. FID is used as the primary metric for evaluating generation quality.

For all experiments, we follow the training configurations used in prior work~\cite{zhai2024normalizing,gu2025starflow}. We employ a cosine learning rate schedule, warming up from $10^{-6}$ to $10^{-4}$ during the first epoch and decaying back to $10^{-6}$ thereafter. A weight decay of $10^{-4}$ is applied for stability. All models are trained end-to-end using the AdamW optimizer with momentum coefficients $(0.9, 0.95)$. All experiments employ Fourier time conditioning~\cite{song2020score}, and we additionally embed the log-scaled timestep $\log t$ following the design of EDM~\cite{karras2022elucidating}.

For pixel-space experiments, the patch size is fixed to $1/32$ of the input resolution, ensuring a constant sequence length across ImageNet-64, 128, and 256. We vary only the channel width while keeping four causal Transformer layers ($L=4$) for all configurations. We adopt the \emph{deep-shallow} architecture from~\citet{gu2025starflow}, summarized in Tab.~\ref{Table:network_arch}.

For latent-space experiments, we use the Big Model in Tab.~\ref{Table:network_arch} with a patch size of~1, consistent with the original STARFlow. We use SD-VAE~\cite{rombach2022high} as the encoder and decoder, following STARFlow.

We set the maximum noise level to $\tmax = 0.3$ for ImageNet-64, $\tmax = 0.5$ for ImageNet-128, and $\tmax = 0.7$ for ImageNet-256 in pixel experiments. For latent-space ImageNet-256 experiments, we use $\tmax = 0.5$. 

For all ablation studies, we use the Small model for pixel space and the Big model for latent space.

\subsection{Quantitative Performance}

We compare iTARFlow with representative diffusion and autoregressive generative models across multiple resolutions. For the qualitative evaluation, please see Fig.~\ref{fig:teaser}.
\paragraph{ImageNet-64.}
Tab.~\ref{tab:imagenet64} reports the FID scores for the Small and Big variants. Relative to TARFlow~\cite{zhai2024normalizing}, iTARFlow achieves better performance with less than half the number of parameters. Scaling the model to 770M parameters results in an FID of 1.68, approaching diffusion-based performance despite using a fundamentally different modeling paradigm.

\paragraph{ImageNet-128.}
Tab.~\ref{tab:imagenet128} shows that iTARFlow achieves competitive performance compared with diffusion models. In comparison to TARFlow, iTARFlow yields substantially improved results while requiring fewer parameters.

\paragraph{ImageNet-256.}
As shown in Tab.~\ref{tab:imagenet256}, iTARFlow improves upon both TARFlow~\cite{zhai2024normalizing} and STARFlow~\cite{gu2025starflow}, and surpasses recent autoregressive normalizing flow models~\cite{zheng2025farmer,li2025fractal}. Notably, the performance gap between normalizing flows and diffusion models is further reduced. As discussed in Sec.~\ref{sec:failure}, part of the remaining gap is attributable to a small number of collapsed samples used in FID evaluation. We expect that resolving these artifacts will further enhance the quantitative performance of iTARFlow.
\begin{table}[t]
  \centering
  \scriptsize
  \renewcommand{\arraystretch}{0.95}
  \caption{Comparison of TARFlow and iTARFlow denoising strategies from noise level $t = 0.3$. The results show that replacing the original TARFlow self-denoiser with our proposed iterative denoiser significantly improves FID, and the iTARFlow model with iterative denoising achieves similar performance on ImageNet-64.}\label{table:iter_ablation}
  \resizebox{\columnwidth}{!}{%
  \begin{tabular}{lc}
  \toprule
  \textbf{Model} & \textbf{FID}  \\
  \midrule
  TARFlow-S @ $t=0.3$ w/ \emph{self denoiser} & 5.22 \\
  TARFlow-S @ $t=0.3$ w/ iTARFlow-S(ours) \emph{iterative denoiser} & 2.08 \\
  iTARFlow-S (ours) @ $t=0.3$ w/ \emph{iterative denoiser} & 2.05 \\
  \bottomrule
  \end{tabular}%
  }
\end{table}

\begin{table}[t]
  \centering
  \scriptsize
  \renewcommand{\arraystretch}{0.95}
  \caption{Comparison of StarFlow and iTARFlow variants evaluated at noise level $t = 0.5$ under different denoising strategies in latent space ImageNet 256 experiment. iTARFlow exhibits comparable performance when paired with either its own iterative denoiser or a DiT-style iterative denoiser. This indicates that the score functions estimated by iTARFlow are on par with those of DiT, despite the fundamentally different training objectives.}\label{Table:DiT_compare}
  \resizebox{\columnwidth}{!}{%
  \begin{tabular}{lc}
  \toprule
  \textbf{Model} & \textbf{FID}  \\
  \midrule
  STARFlow-B @ $t=0.3$ w/ \emph{finetune decoder} & 2.48 \\
  STARFlow-B @ $t=0.5$ w/ \emph{DiT iterative denoiser}  & 2.54 \\
  iTARFlow-B (ours) @ $t=0.5$ w/ \emph{iterative denoiser} & 2.32 \\
  iTARFlow-B (ours) @ $t=0.5$ w/ \emph{DiT iterative denoiser} & 2.28 \\
  \bottomrule
  \end{tabular}%
  }
\end{table}

\begin{figure}[t]  % use [t] or [b] to control placement
  \centering
  \includegraphics[width=\columnwidth]{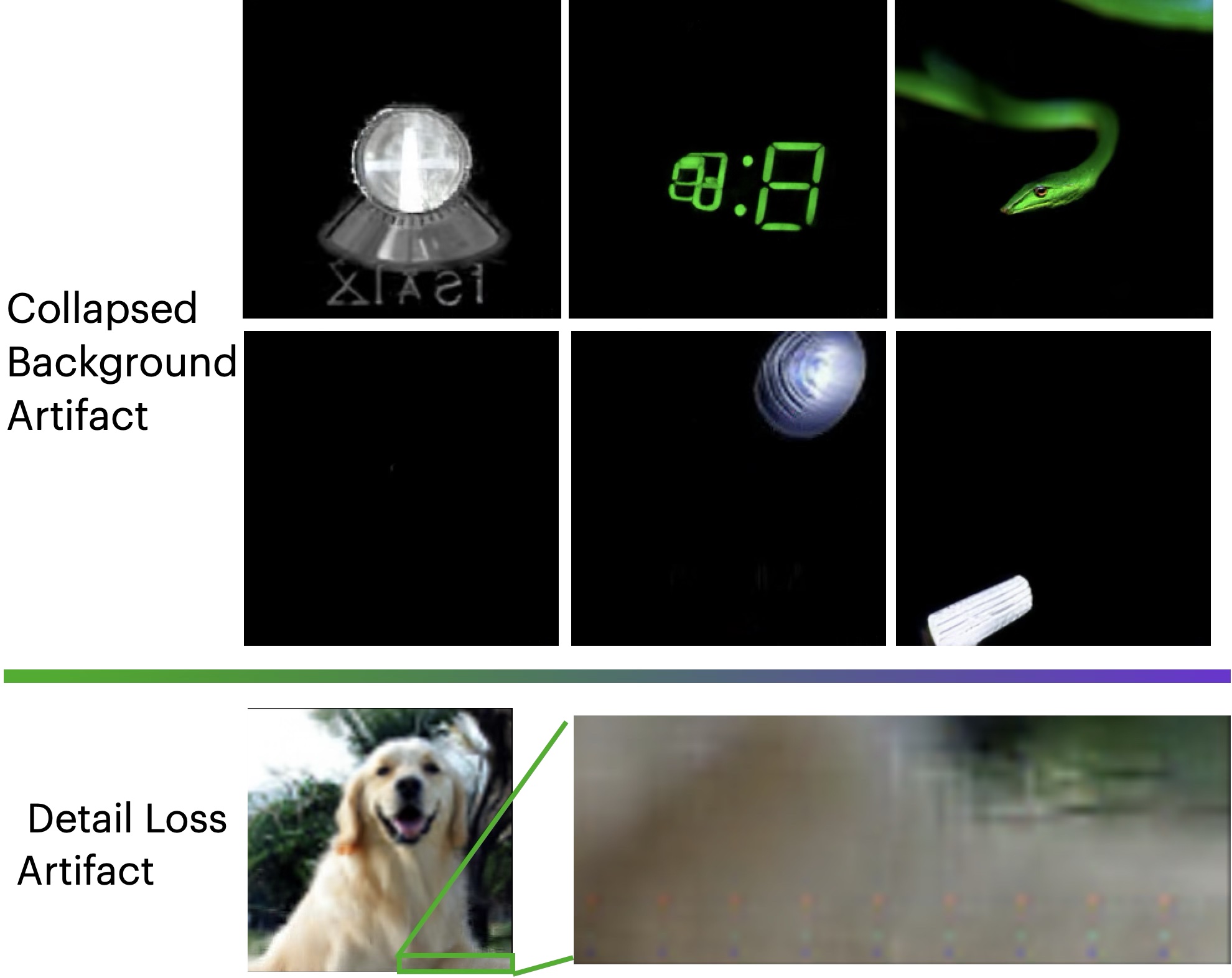}
  \caption{Two primary artifact types are observed in the 50k iTARFlow samples (FID = 3.53). (a) A collapsed-background artifact, where the model produces an entirely black background. (b) A low-frequency blur artifact, characterized by an increased proportion of blurry images compared to the original training set.}
  \label{fig:failure_case}
  \vspace{-10pt}
\end{figure}
\subsection{Analysis of Score Estimation}

Tab.~\ref{table:iter_ablation} provides an ablation study of the iterative denoising behavior learned by iTARFlow. When using a self-denoiser, generation quality degrades significantly under high training noise, consistent with the visualization in Fig.~\ref{fig:noise_dilimma}. However, this degradation is mitigated when denoising a TARFlow model using an \emph{independently} trained iTARFlow model. In this case, the performance becomes comparable to iTARFlow, demonstrating that the score estimated by iTARFlow generalizes well and functions effectively as a universal denoiser for general TARFlow.

We further compare the iTARFlow score estimator with DiT~\cite{peebles2023scalable} in Tab.~\ref{Table:DiT_compare}.  Prior work~\cite{gu2025starflow} reported that incorporating DiT denoisers into STARFlow often degrades performance compared with a fine-tuned VAE,
a trend that we also observe in our experiments.  In contrast, iTARFlow, which benefits from amortized noise-level training and iterative denoising, outperforms
STARFlow combined with a standard fine-tuned VAE. Interestingly, the samples produced by iTARFlow's iterative denoiser are comparable in quality to those denoised
directly by a DiT model, indicating that the score estimated by iTARFlow achieves a similar level of accuracy as DiT even though the parameterization is significantly different. In addition, iTARFlow is markedly more training-efficient, converging in far fewer epochs than DiT (600 vs.\ 1400), consistent with prior findings in STARFlow. More importantly, unlike STARFlow, iTARFlow integrates naturally with DiT-based denoisers without any observable performance degradation, indicating a promising direction toward a unified hybrid autoregressive diffusion framework for continuous tokens.

\subsection{Analysis of Generation Failure}\label{sec:failure}

Although iTARFlow achieves a competitive FID of 3.32 on ImageNet-256 in the pixel space, noticeable failure cases remain. In the 50k generated samples used for FID computation, we observe two recurring issues: (1) collapsed backgrounds, where the entire background becomes black, and (2) a higher proportion of blurry images compared to the training set. Importantly, most blurry samples share a consistent structure: the bottom-right region, which corresponds to the first continuous token generated autoregressively, exhibits a characteristic distortion (Fig.~\ref{fig:failure_case}). For the collapsed background artifact, we hypothesize that it arises from the increased CFG scale used in the large-patch experiments, as shown in Fig.~\ref{fig:cfg_ablation}. This issue could potentially be mitigated by designing a more principled and concise guidance design during generation. A plausible explanation for detail loss artifacts is that the first autoregressive token is poorly aligned with the training distribution, since it is produced without any conditioning context. This lack of preceding tokens often results in an out-of-distribution initial prediction, which subsequently propagates errors to later tokens in the sequence. This effect is amplified as patch dimensionality increases, explaining the elevated failure rate at ImageNet-256 with patch size~8. Addressing these two artifacts represents a promising direction for further improving high-resolution performance.

%% file: subtex/conclusion.tex
\section{Conclusion and Limitations}
% In this paper, we introduced iTarFlow, which is a simple but effect extension of previous TarFlow. By training the TarFlow with different noise level, this atuoamtically allow us to conduct iterative denosing from varying noise level to overcome noise dillema phenomenon (Fig.\ref{fig:noise_dilimma}) in normalzing flow. By doing so, we largely improve the performance of normalizing flow in the different experiment setting which narrowing the gap compared with SOTA model on the market. We also analyze the observation which prevent iTarFlow to further improve, which may pave the way for further improvement. Furthermore, the TarFlow branch of work used to fail to integrate with existing diffusion model which makes it less flexible, which is different from previous AR+Diffusion framework. However, in Tab.\ref{Table:DiT_compare}, we demonstrates that iTarFlow is indeed can be able to be integrate with pretrained DiT. We spectaculr taht this is due to the amortize trainnig with different noise level which is the only different variable in the comparsion setting in Tab.\ref{Table:DiT_compare} which return such flexiblity to the atuoregressive normalzing Flow.

% In terms of limitation is also obvious.

% \section{Conclusion}

In this paper, we introduced iTARFlow, a simple yet effective extension of the TARFlow framework. 
By training the autoregressive normalizing flow across a spectrum of noise levels, iTARFlow naturally enables iterative denoising from coarse to fine noise scales, effectively mitigating the \emph{noise dilemma} phenomenon (Fig.~\ref{fig:noise_dilimma}) commonly observed in normalizing flows. 
This design significantly improves generation quality across multiple experimental settings, substantially narrowing the performance gap between normalizing flows and state-of-the-art generative models.
We also investigated the primary limitations that currently prevent iTARFlow from achieving even stronger performance, offering insights that may inspire future improvements.
Furthermore, prior TARFlow variants \cite{gu2025starflow} struggled to integrate with pretrained diffusion backbones, reducing their flexibility compared with existing AR+Diffusion hybrid frameworks \cite{yu2022scaling,ramesh2022hierarchical}. 
In contrast, our results in Tab.~\ref{Table:DiT_compare} demonstrate that iTARFlow can be seamlessly combined with pretrained DiT models. 
We hypothesize that this improved compatibility arises from the amortized training across noise levels—the key distinguishing factor in the comparison, which restores flexibility to the NF.

In terms of limitations, our experiments indicate that one can still expect a quality gap compared to the current SOTA. Automatic differentiation during iterative denoising consumes significant GPU memory, which is expected since it applies autograd to the entire network. This is the major computational limitation of our method. On the theory front, there is currently no known theory for finding the optimal hyperparameters mentioned in Sec.~\ref{sec:train}, which creates a larger room for tuning.

\section{Impact Statement}
This paper concerns fundamental techniques for generative modeling. There are many potential societal
consequences of our work, none which we feel must be
specifically highlighted here.